\documentclass[runningheads]{llncs}
\usepackage{graphicx}
\usepackage{subfigure}
\usepackage{algorithm} %format of the algorithm
\usepackage{algorithmic} %format of the algorithm
\usepackage{multirow}
\usepackage[table]{xcolor}
\usepackage{bbding}

\begin{document}

\definecolor{red}{rgb}{1.0,0,0}
\definecolor{green}{rgb}{0.0,1,0}
\definecolor{blue}{rgb}{0,0.7,1}
\definecolor{gray}{rgb}{0.5,0.5,0.6}
\definecolor{brown}{rgb}{0.5,0.2,0.2}
\definecolor{yellow}{rgb}{1,1,0}
\definecolor{RosyBrown}{rgb}{0.7,0.56,0.1}
\definecolor{SandyBrown}{rgb}{0.95,0.64,0.37}
\definecolor{Cornsilk2}{rgb}{0.93,0.91,0.79}

\title{Joint Multi-frame Detection and Segmentation for Multi-cell Tracking}
%
%\titlerunning{Abbreviated paper title}
% If the paper title is too long for the running head, you can set
% an abbreviated paper title here
%
\author{Zibin Zhou \and
Fei Wang \and
Wenjuan Xi \and
Huaying Chen \\
Peng Gao \and
Chengkang He}
\authorrunning{Z. Zhou, F. Wang, et al.}
% First names are abbreviated in the running head.
% If there are more than two authors, 'et al.' is used.
%
\institute{Harbin Institute of Technology, Shenzhen
%\email{\{zhouzibin, xiwenjuan, pgao, hechengkang\}@stu.hit.edu.cn}\\
%\email{\{wangfeiz, chenhuaying\}@hit.edu.cn}
}
\maketitle              % typeset the header of the contribution
\begin{abstract}
Tracking living cells in video sequence is difficult, because of cell morphology and high similarities between cells. Tracking-by-detection methods are widely used in multi-cell tracking. We perform multi-cell tracking based on the cell centroid detection, and the performance of the detector has high impact on tracking performance. In this paper, UNet is utilized to extract inter-frame and intra-frame spatio-temporal information of cells. Detection performance of cells in mitotic phase is improved by multi-frame input. Good detection results facilitate multi-cell tracking. A mitosis detection algorithm is proposed to detect cell mitosis and the cell lineage is built up. Another UNet is utilized to acquire primary segmentation. Jointly using detection and primary segmentation, cells can be fine segmented in highly dense cell population. Experiments are conducted to evaluate the effectiveness of our method, and results show its state-of-the-art performance.

\end{abstract}
\section{Introduction}

Multi-cell tracking in image sequence is valuable for stem cell research, tissue engineering, drug discovery and proteomics~\cite{ref_article1}. Researchers can construct cell lineage trees and analyze cell morphology based on cell tracking results~\cite{ref_article2}.

Cell tracking is more challenging than general object tracking. Firstly, cells may be deformed, such as elongation, contraction, and swelling~\cite{ref_article3}. Secondly, there is a very high similarity between cells. Cells of same kind have the same internal structure, and they are difficult to be distinguished through their appearance. In addition, the irregular motion of cells, the mitotic behavior, the complexity of the background, and the interference of other impurities also increase the challenge.

The development of deep learning in recent years has greatly promoted the progress in computer vision. For example, the performance of ResNet~\cite{ref_article4} exceeded the performance of humans on the ImageNet test set. Cell tracking has evolved from contour evolution, filtering templates, to tracking-by-detection methods~\cite{ref_article2}. Researchers continue to improve the robustness of algorithms.

There is no such a multi-cell tracking algorithm that can perform well in all varieties of video sequences. For example, Ref.~\cite{ref_article5} can perform segmentation and tracking very well when cells are large, but not well when cells are small. Highly dense small cells are apt to be missed during tracking, and lead to tracking errors. To solve this challenge, we propose an algorithm that jointly using detection and segmentation for multi-cell tracking. Our method is composed of four portions: cell centroid detection with multi-frame images, primary multi-cell tracker, primary cell segmentation, fine segmentation. Each portion will be detailed in Sect.~\ref{sect_Method}. The contributions of our work can be summarized as follows:
\begin{itemize}
\item[$\bullet$] Multi-frame as input to UNet~\cite{ref_article6} is proposed, which helps the network to extract spatio-temporal information. Detection performance of mitotic cells is improved, therefore detection performance of mitosis is improved by the mitosis detection algorithm when tracking.
\item[$\bullet$] A fine cell segmentation algorithm is proposed for tracking highly dense small cells. By jointly using primary tracking results of cell centroid detection and primary cell segmentation results, we achieve a new state-of-the-art performance on dataset Fluo-Hela~\cite{ref_article7}.
\end{itemize}
\noindent Effectiveness of our method is evaluated with Cell Tracking Benchmark~\cite{ref_article7} of Cell Tracking Challenge (www.celltrackingchallenge.net). Performance metrics include tracking accuracy, segmentation accuracy, and a combination performance of both.

\section{Related works}
Tracking-by-detection methods are widely used in multi-object tracking~\cite{ref_article8,ref_article9}, as well as in multi-cell tracking. Starting with detecting or segmenting cells in a video sequence, these methods establish temporal associations for cells in frames to frames. Detection performance has high impact on tracking performance [10]. As long as good detection results are available, the tracking problem can be simplified~\cite{ref_article8}. This paper proposes a tracking-by-detection method, which focuses on detection and
segmentation. Detection or segmentation in tracking-by-detection methods will be briefly reviewed as below.

Ciresan et al.~\cite{ref_article11} proposed to use neural networks for segmentation of microbial images in the early days. They use a neural network as a pixel classifier to segment the biological neuron membrane. The network inference must be run on patch-by-patch separately. Unfortunately patches overlap each other, a large amount of computation redundancy occurs, therefore calculation is quite slow.

Ronneberger et al.~\cite{ref_article6} proposed the semantic segmentation network, i.e. UNet, cell segmentation was further developed. The network is of the encoder-decoder structure, its input is first be downsampled and then upsampled. In the upsampling process, low-layer features corresponding to the downsampling layer are connected to the corresponding upsampling layer. The network gets high-level semantics without losing much low-level information and works well with only small amount of training data. Zhou et al.~\cite{ref_article12} redesigned the skip-connection of UNet and improved segmentation performance by reducing semantic differences of feature maps in encoder and decoder subnetworks.

Payer et al.~\cite{ref_article5} integrated ConvGRU into a stacked hourglass network for instance segmentation and tracking of cells. ConvGRU not only extracts local features, but also memorizes inter-frame information~\cite{ref_article13}. The stacked hourglass network is similar to UNet, its input is first be downsampled and then upsampled~\cite{ref_article14}. This integrated structure can perform cell segmentation well even in the case of a very close membrane. However, when cells are small, it does not perform well. Arbelle et al.~\cite{ref_article15} proposed a network structure for inter-frame segmentation combining ConvLSTM and UNet. Like ConvGRU, ConvLSTM has spatio-temporal characteristics~\cite{ref_article16}. The integrated structure can perform excellent segmentation even in the case of partial disappearance of cells, but does not perform well in the case of less training data. The work of Payer et al.~\cite{ref_article5} and Arbelle et al.~\cite{ref_article15} are similar.

As long as cells are accurately detected or segmented, tracking will be largely simplified. In Ref.~\cite{ref_article5}, although only the intersection over union is used for inter-frame cell associations, it achieves desired tracking performance.

\begin{figure}
	\centering
	\includegraphics[width=0.9\linewidth]{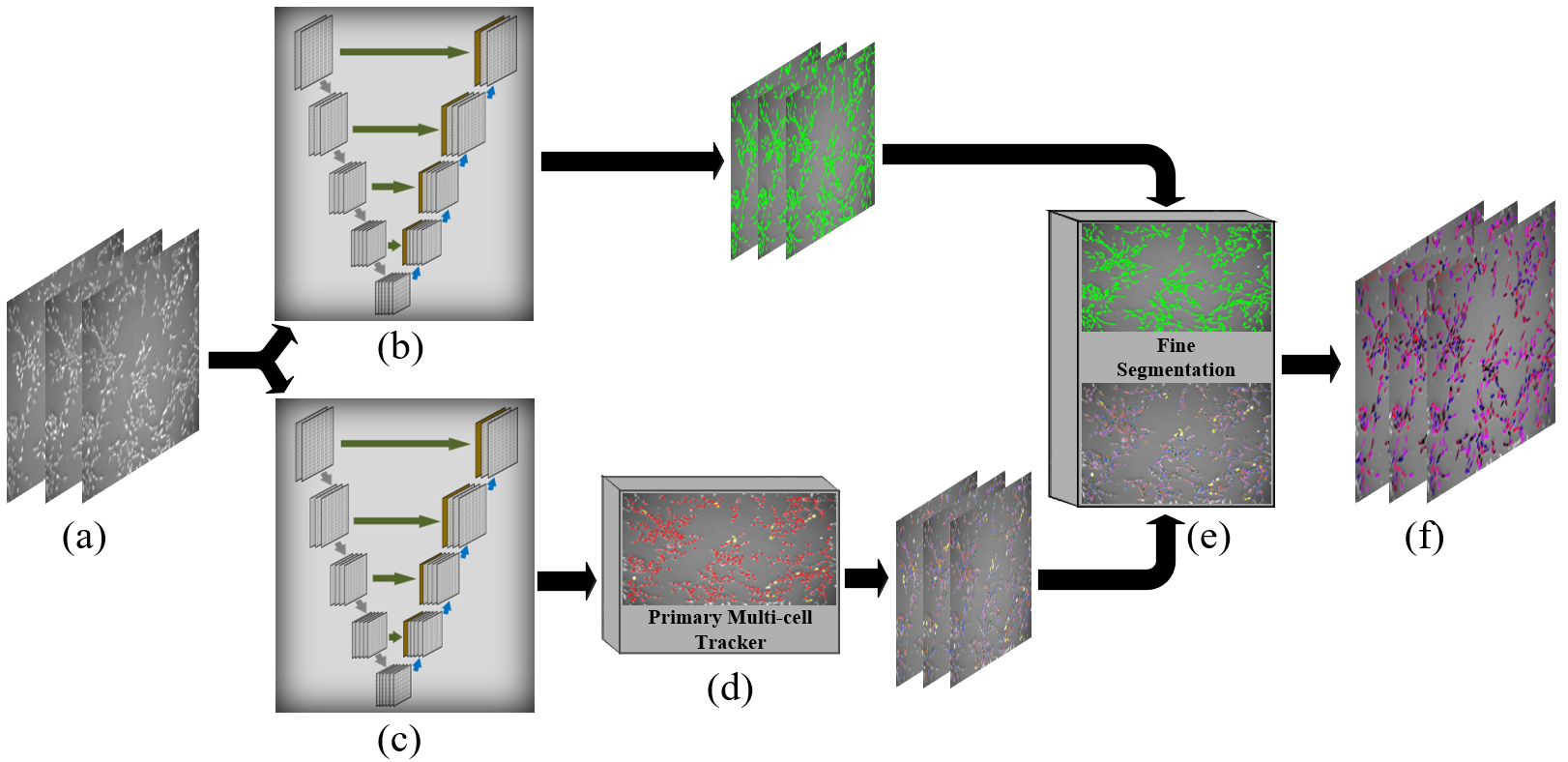}
	\caption{Overview of our proposed tracking framework. (a) Input. (b) UNet for primary cell segmentation. (c) UNet for cell centroid detection with multi-frame images. (d) Primary multi-cell tracker. (e) Fine segmentation. (f) Final tracking results.}
	\label{fig_frame}
\end{figure}

\section{Method} \label{sect_Method}
In this section our proposed method is detailed. As shown in Fig.~\ref{fig_frame}, it has four portions: cell centroid detection with multi-frame images, primary multi-cell tracker, primary cell segmentation, fine segmentation.

\subsection{Cell Centroid Detection with Multi-frame Images}  \label{sect_Detection}
To identify cells in highly dense population, it is a useful technique to identify cell centroid first. Here UNet~\cite{ref_article6} (UNet-DET) is used to locate cell centroid.

\begin{figure}[htbp]
    \centering
    \subfigure[]{
    \includegraphics[width=0.25\linewidth]{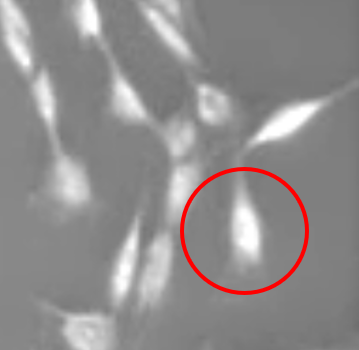} }
    \subfigure[]{
    \includegraphics[width=0.25\linewidth]{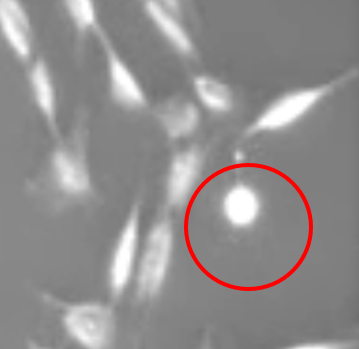} }
    \subfigure[]{
    \includegraphics[width=0.25\linewidth]{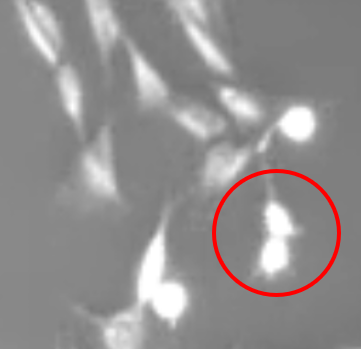} }
    \caption{Morphological changes in mitosis.}
    \label{fig_mitosis}
\end{figure}

Mitotic cells are defined as those cells before, during and after mitosis. During mitosis, obvious morphological changes usually occur, which make them look different from normal cells, i.e., cells in non-mitosis status. In Fig.~\ref{fig_mitosis}, from left to right, morphological changes of a cell before, during and after mitosis.

\noindent Pixels are categorized into three categories: mitotic cells, normal cells and backgrounds. If information in previous nearby frames is included, network can more accurately learn to identify mitotic cells~\cite{ref_article17}.

Different from usual single-frame input method, we feed incorporative consecutive pre-\emph{N$_{input}$} frames into the network. This approach does improve cell centroid detection a lot. Combined with past image information, the network can extract living cell behavioral features, and then screen out impurities that do not change shape.

Compared with Ref.~\cite{ref_article5,ref_article15}, the overall network complexity based on UNet-DET does not increase much. Only the number of weight parameters in the first layer of the network increases.

A cross-entropy loss function is used to train the network. Mitotic cells are few, and more attention should be paid to them by setting a weight map. The network loss function is defined as in Ref.~\cite{ref_article6}:

\begin{equation}
L = {\rm{ - }}\frac{1}{T}\sum\limits_{i = 1}^T {w(i)\log \frac{{\exp (h(i,g(i)))}}{{\sum\limits_{j = 1}^C {\exp (h(i,j))} }}}
\end{equation}

\noindent where \emph{T} denotes the total number of pixels. \emph{w($\cdot$)} is a weight map. \emph{g($\cdot$)} is the real category corresponding to the input pixel. \emph{h(i,j)} denotes the final output of category \emph{j} when input pixel \emph{i}.

Thresholding inference results, using the flood-fill algorithm~\cite{ref_article19} to fill internal holes of connected area. Extracting contour of connected area, computing its centroid as cell centroid. High-precision cell centroid detection and classification are acquired without increasing much network overhead.

\subsection{Primary Multi-cell Tracker} \label{sect_Tracking}
Different from common object tracking, cell lineage needs to be built up during tracking. If a cell is categorized as a mitotic cell in multiple consecutive frames, it is highly likely to undergo mitosis. A mitosis detection algorithm, using a local cell status matrix, is proposed. The matrix is created in the beginning of new trajectory. Elements of the matrix record the status of a cell: the centroid position (X, Y), its sequence number (Z), whether it is a normal or mitotic cell. When the number of mitotic cells is larger than a given threshold, it can be concluded that mitosis occurs.

As long as images are captured in high frame rate and cells are precisely detected, it is possible to use overlap intersection-over-union (IOU) to build inter-frame object associations ~\cite{ref_article8,ref_article18}, and then ideal tracking is performed.

When the cell centroid is located, a bounding box with a size of \emph{N$_{size}$$\times$N$_{size}$} is created around the centroid, where \emph{N$_{size}$} is the average length of cells scaled in number of pixels. With the assumption that sequences have the high enough frame rate, inter-frame cells can be associated by only using the IOU of bounding boxes. Each newly detected cell is needed to be associated with an existing trajectory. The association strategy is computed as:

\begin{equation}
\Theta (t,D) = \mathop {arg{\mathop{\rm ma}\nolimits} }\limits_{d \in D} x(\Lambda (d,t))\
\end{equation}
\begin{equation}
d(t,D) = \left\{ {\begin{array}{*{20}{c}}
{\Theta (t,D),}\\
{N,}
\end{array}} \right.\begin{array}{*{20}{c}}
{\Lambda (\Theta (t,D),t) \ge \alpha }\\
{\Lambda (\Theta (t,D),t) < \alpha }
\end{array}
\end{equation}

\noindent where \emph{t} denotes the cell at the end of the trajectory to be associated with, \emph{D} the set of candidate detected cells in the current frame, \emph{$\Lambda$($\cdot$,$\cdot$)} the IOU of both. \emph{$\alpha$} is the minimum overlapping intersection that allows to associate. \emph{N} denotes that there are no associating candidate cells.

If the largest IOU that is less than \emph{$\alpha$}, the candidate cell will be discarded.

The trajectory is terminated when it has no associated cells. For a candidate cell that has no existing trajectory to associate, a temporary new trajectory is created. Short trajectories are most likely pseudo trajectories caused by impurity interference, and therefore are discarded.

If a mitosis event is detected, the original trajectory is terminated. At the same time, cell lineage is established between the mitotic mother cell and its two newborn daughter cells. This distinguishes normal cells which newly enter the field of view from newborn daughter cells.

Primary tracking results of cell centroid detection are acquired, cell IDs of the same trajectory are same.

\subsection{Primary Cell Segmentation}  \label{sect_Segmentation}
UNet~\cite{ref_article6} (UNet-SEG) is used for primary segmentation of cells. During this stage, image pixels are categorized into cell boundaries, cell interiors, and backgrounds. The cross-entropy loss function is used to train the network. More attention is paid on cell boundaries by setting a weight map.

Threshold is performed to acquire cell segmentation by using inference results of cell boundaries and cell interiors. There may be holes inside segmented cells and the flood-fill algorithm~\cite{ref_article19} is used to fill these internal holes.

\begin{figure}[htbp]
    \centering
    \subfigure[]{
    \includegraphics[width=0.3\linewidth]{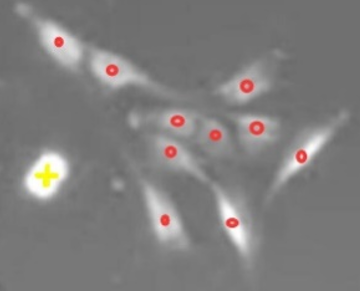} }
    \subfigure[]{
    \includegraphics[width=0.3\linewidth]{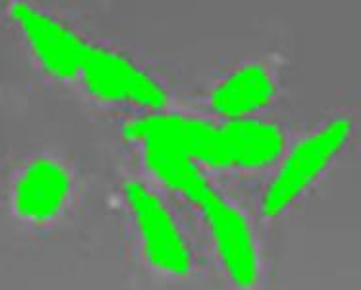} }
    \subfigure[]{
    \includegraphics[width=0.3\linewidth]{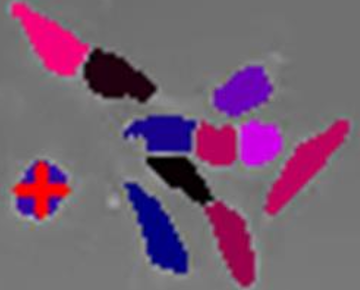} }
    \caption{Dense cell segmentation results. Cross: mitotic cells. Dot: normal cells. (a) Original image and cell centroid detection results. (b) Primary cell segmentation results. (c) Fine segmentation results.}
    \label{fig_seg}
\end{figure}

When many cells are close each other, each cell is not separated at this stage. When primary segmentation is finished, as shown in Fig.~\ref{fig_seg} (b), cells close each other may be segmented as a blob, a piece of connected area.

\subsection{Fine Segmentation}
Results from primary segmentation may contain many connected area as shown in Fig.~\ref{fig_seg} (b). When cells are dense, multiple cells are segmented together. In this section, fine segmentation is conducted to separate each cell individually, which jointly use primary tracking results of cell centroid detection from Sect.~\ref{sect_Tracking} and primary cell segmentation results from Sect.~\ref{sect_Segmentation}.

Assuming cell boundary is closest to its centroid for non-overlapping small cells of similar size, each pixel in a connected area is assigned to a cell centroid contained in this connected area according to the following formulation:

\begin{equation}
P({p_{pixel}}) = \mathop {\arg \min }\limits_{p \in {P_{\det }}} (d({p_{pixel}},p))
\end{equation}

\noindent where \emph{P$_{det}$} is a set of cell centroid contained in a connected area, and \emph{p$_{pixel}$} denotes pixels in the connected area, \emph{d($\cdot$,$\cdot$)} denotes the Euclidean distance.

However, it would be a very time-consuming task to calculate Euclidean distance from each pixel to each cell centroid. Therefore, Voronoi~\cite{ref_article20} is used to accelerate pixel assignment. Pixel assignment is shown as Algorithm~\ref{alg_seg}. Fig.~\ref{fig_seg} (c) shows fine segmentation results.

\begin{algorithm}[htb]
  \caption{Fine Segmentation}
  \label{alg_seg}
  \begin{algorithmic}[1]
    \REQUIRE   ~~\\
    Primary cell segmentation \emph{SEG = {SEG$_{1}$, SEG$_{2}$, …, SEG$_{M}$}}; \\
    Primary tracking results of cell centroid detection \emph{P$_{det}$ = {P$_{1}$, P$_{2}$, …, P$_{N}$}};
    \ENSURE  ~~\\
    Fine segmentation result \emph{SEG\_RES};
    \STATE Initialize a zero matrix \emph{SEG\_RES};
    \STATE Calculate the Voronoi diagram according to \emph{P$_{det}$};
    \STATE Label the connected domain of Voronoi diagram, acquire \emph{LAB\_VOR};
    \FOR {\emph{seg} in \emph{SEG}}
        \STATE \emph{P$_{inside}$} = cell centroid contained by \emph{seg};
          \IF {len(\emph{P$_{inside}$})==1}
            \STATE \emph{SEG\_RES} += \emph{seg};
          \ELSE
            \IF{len(\emph{P$_{inside}$)}$>$1}
                \FOR {\emph{p} in \emph{P$_{inside}$}}
                    \STATE \emph{SEG\_RES} += \emph{seg} $\bigcap$ \emph{LAB\_VOR(p)};
                \ENDFOR
            \ENDIF
          \ENDIF
    \ENDFOR
  \end{algorithmic}
\end{algorithm}

\section{Experiment}
The effectiveness of our method is evaluated with datasets of Cell Tracking Challenge~\cite{ref_article9}. The dataset for each kind of cells provides two training sequences with GT (ground truth) and two test sequences without GT. GT includes TRA (tracking) GT and SEG (segmentation) GT. TRA GT essentially contains cell centroid of all sequences. SEG GT is few, which makes cell segmentation more difficult.

\subsection{Training}
UNet-DET and UNet-SEG both are composed of 5 downsampling layers and 5 upsampling layers. Adam optimizer~\cite{ref_article21} is used and the learning rate is set as \emph{0.001}. The exponential decay rate of learning rate is set as \emph{0.95}, the global step size of the decay is set as \emph{4} times of the number of sample set. A weighted cross-entropy loss function is used for training to pay more attention on mitotic cells or cell boundaries. To augment samples, we apply horizontal and vertical flips, and randomly add Gaussian noise or salt and pepper noise on sample set.

For UNet-DET, incorporative consecutive pre-\emph{N$_{input}$} frames are fed into the network. The dimension of input is (\emph{H}, \emph{W}, \emph{N$_{input}$)}, here \emph{N$_{input}$ = 3}. The label is TRA GT of last frame. We have modified TRA GT. Mitotic cells are defined as cells of \emph{N$_{mitisis}$} frames before and after mitosis, here \emph{N$_{mitosis}$ = 2}. Weight of each category is set as: \emph{0.5}-mitotic cells, \emph{0.3}-normal cells, \emph{0.2}-backgrounds.

For UNet-SEG, SEG GT are categorized into cell boundaries and cell interiors. When the amount of SEG GT is scarce, original images and SEG GT are cropped centered on each cell centroid. \emph{N$_{cell}$} training samples with a size of \emph{S$_{crop}$$\times$S$_{ceop}$} are cropped from each original image. Here \emph{S$_{crop}$} is set as \emph{5} times of the mean size of cells, \emph{N$_{cell}$} denotes the number of cells in the image. Weight of each category is set as: \emph{0.5}-cell boundaries, \emph{0.3}-cell interiors, \emph{0.2}-backgrounds.

\subsection{Comparison of Multi-frame Input and Single-frame Input}
The advantage of multi-frame input over single-frame input is evaluated with Cell Tracking Challenge datasets PhC-PSC and Fluo-HeLa~\cite{ref_article7}, which have more mitosis. The first half of the two training sequences are used as training samples and the other half are used as test samples. Performance of cell centroid detection is scaled with three metrics, i.e., Precision, Recall, F1-score. These metrics on normal cells and mitotic cells are shown in Table~\ref{tab_detection} respectively.

From Table~\ref{tab_detection}, three metrics are improved 15\% with mitotic cells, while slightly improved with normal cells. Inter-frame morphological changes of mitotic cells are obvious, retaining the historical information can improve detection performance.

In our multi-cell tracking algorithm, the improved mitotic cells detection performance will improve the mitosis event detection performance. Table~\ref{tab_mitosis} shows the mitosis event detection performance compared with different input modes. Due to our mitosis detection algorithm is strict to determine whether mitosis has occurred, Precision is high of both. The other two metrics of multi-frame input are improved about 20\%. Therefore, the reconstruction of the cell lineage will be improved.

\begin{table}
\caption{Detection performance of normal cells and mitotic cells}\label{tab_detection}
\centering
\begin{tabular}{c|c|c|c|c}
\hline
Status & Input Frame & Precision (\%) & Recall (\%) & F1 score (\%) \\
\hline
\multirow{2}{*}{Mitotic cells} &  1 & 44.3 & 49.7 & 46.8 \\ \cline{2-5}
&  3 & {\bfseries 60.1} & {\bfseries 65.3} & {\bfseries 62.6} \\
\hline
\multirow{2}{*}{Normal cells} &  1 & 93.2 & 91.6 & 92.4 \\ \cline{2-5}
&  3 & {\bfseries 94.3} & {\bfseries 93.7} & {\bfseries 94.0} \\
\hline
\end{tabular}
\end{table}

\begin{table}
\caption{Detection performance of mitosis event}\label{tab_mitosis}
\centering
\begin{tabular}{c|c|c|c|c}
\hline
Dataset & Input Frame & Precision (\%) & Recall (\%) & F1 score (\%) \\
\hline
\multirow{2}{*}{Phc-PSC} &  1 & {\bfseries 87.6} & 26.8 & 41.1 \\ \cline{2-5}
&  3 & 86.0 & {\bfseries 47.8} & {\bfseries 61.5} \\
\hline
\multirow{2}{*}{Fluo-Hela} &  1 & 77.3 & 37.2 & 50.2 \\ \cline{2-5}
&  3 & {\bfseries 87.7} & {\bfseries 63.6} & {\bfseries 73.7} \\
\hline
\end{tabular}
\end{table}

\subsection{Evaluations on Cell Tracking Benchmark}
Cell tracking performance is evaluated in Cell Tracking Benchmark with 2D datasets of Cell Tracking Challenge~\cite{ref_article7}. Performance of tracking is scaled in TRA (tracking accuracy), SEG (segmentation accuracy), and OP$_{CTB}$ (the mean of both)~\cite{ref_article7}.

Based on the ranking announced on the day April 30th, 2019, which can be seen at the web of Cell Tracking Challenge~\cite{ref_article9}, performance metrics of our proposed method are listed in Table~\ref{tab_tracking}. For each dataset, generally performance metrics of more than 20 methods are ranked~\cite{ref_article9}, including the original UNet (FR-Ro-GE)~\cite{ref_article6}, the globally trained UNet (FR-Fa-GE)~\cite{ref_article22}, the method of ConvLSTM integrated into UNet (BGU-IL)~\cite{ref_article15}, the method of ConvGRU integrated into the stacked hourglass network (TUG-AT)~\cite{ref_article5}, and the method of global threshold and global associations with spatio-temporal (KTH-SE)~\cite{ref_article9}.

\begin{table} [htbp]
\caption{Quantitative Comparison of Cell Tracking Benchmark (\%)}\label{tab_tracking}
\centering
\begin{tabular}{c|c|c|c|c|c|c c}

\hline
 &   & Phc-PSC & Fluo-Hela & Fluo-SIM+ & Fluo-GOWT1 & & \multirow{2}{*}{}  \\
\hline

\multirow{4}{*}{OP$_{CTB}$} & 1$^{st}$ &  \cellcolor{blue}80.4 & \cellcolor{red}95.3 & \cellcolor{SandyBrown}88.2 & \cellcolor{green}95.1 & & \\ \cline{2-6}
& 2$^{nd}$ & \cellcolor{gray}80.4 & \cellcolor{gray}94.4 & \cellcolor{gray}88.1 & \cellcolor{brown}93.4 & & \cellcolor{red}Ours \\ \cline{2-6}
& 3$^{rd}$ & \cellcolor{red}80.1 & \cellcolor{green}94.2 & \cellcolor{yellow}87.8 & \cellcolor{Cornsilk2}92.3 & & \cellcolor{green}KTH-SE(1-4) \\ \cline{2-6}
& & & & \cellcolor{red}7$^{th}$ 86.0 & \cellcolor{red}12$^{th}$ 87.0 & & \cellcolor{blue}HD-Hau-GE \\
\hline
\hline

\multirow{4}{*}{SEG} & 1$^{st}$ &  \cellcolor{gray}68.2 & \cellcolor{red}91.9 & \cellcolor{gray}80.7 & \cellcolor{green}92.7 & & \cellcolor{gray}CVUT-CZ \\ \cline{2-6}
& 2$^{nd}$ & \cellcolor{blue}66.5 & \cellcolor{yellow}90.3 & \cellcolor{SandyBrown}80.2 & \cellcolor{brown}92.1 & & \cellcolor{brown}RWTH-GE \\ \cline{2-6}
& 3$^{rd}$ & \cellcolor{red}65.3 & \cellcolor{RosyBrown}90.2 & \cellcolor{green}79.2 & \cellcolor{gray}89.4 & & \cellcolor{yellow}FR-Ro-GE \\ \cline{2-6}
& & & & \cellcolor{red}4$^{th}$ 78.4 & \cellcolor{red}16$^{th}$ 79.0 & & \cellcolor{RosyBrown}FR-Fa-GE \\
\hline
\hline

\multirow{4}{*}{TRA} & 1$^{st}$ &  \cellcolor{green}95.9 & \cellcolor{green}99.1 & \cellcolor{yellow}97.5 & \cellcolor{Cornsilk2}97.9 & & \cellcolor{SandyBrown}BGU-IL(1-4) \\ \cline{2-6}
& 2$^{nd}$ & \cellcolor{red}95.0 & \cellcolor{brown}99.1 & \cellcolor{Cornsilk2}97.3 & \cellcolor{green}97.6 & & \cellcolor{Cornsilk2}TUG-AT \\ \cline{2-6}
& 3$^{rd}$ & \cellcolor{blue}94.3 & \cellcolor{gray}98.8 & \cellcolor{SandyBrown}96.6 & \cellcolor{SandyBrown}96.7 & &  \\ \cline{2-6}
& & & \cellcolor{red}4$^{th}$ 98.7 & \cellcolor{red}9$^{th}$ 93.6 & \cellcolor{red}4$^{th}$ 94.9 & &  \\
\hline
\end{tabular}
\end{table}

\noindent Experiments show excellence of our method on datasets Phc-PSC and Fluo-Hela, highly dense cell datasets. Our method achieves new state-of-the-art performance on SEG and OP$_{CTB}$ of dataset Fluo-Hela. 2$^{nd}$ on TRA of dataset Phc-PSC is achieved. Though these two datasets have very few SEG GT, our method performs excellent.

Our method does not perform very well on datasets Fluo-SIM+ and Fluo-GOWT1. A possible solution to this problem is fully taking advantage of image information when making primary cell segmentation split.

\begin{figure}[htbp]
    \centering
    \subfigure[]{
    \includegraphics[width=0.45\linewidth]{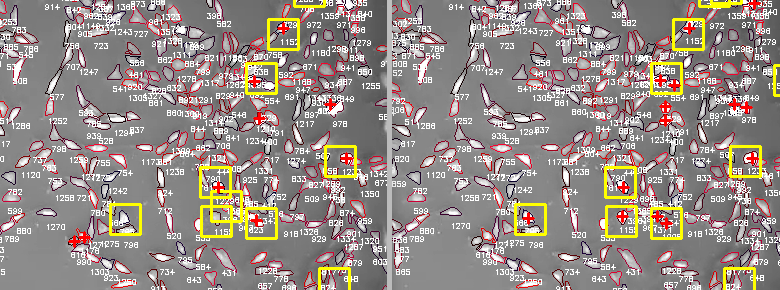} }
    \subfigure[]{
    \includegraphics[width=0.45\linewidth]{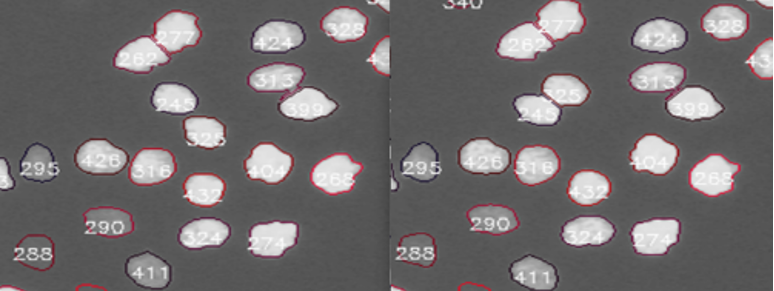} }
    \subfigure[]{
    \includegraphics[width=0.45\linewidth]{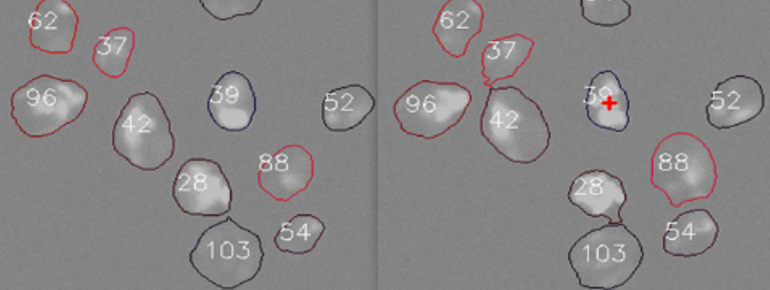} }
    \subfigure[]{
    \includegraphics[width=0.45\linewidth]{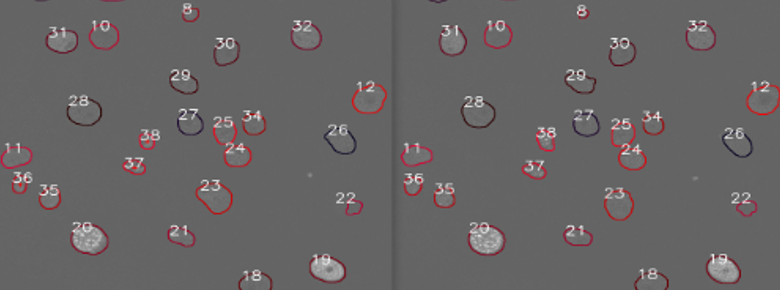} }
    \caption{Cell tracking results. For each pair of images, the left one is the previous frame, and the right one is the current frame. White numbers: trajectory IDs. Yellow boxes: detected mitosis. Red crosses: mitotic cells. Datasets: (a) Phc-PSC. (b) Fluo-Hela. (c) Fluo-SIM+. (d) Fluo-GOWT1.}
    \label{fig_results}
\end{figure}

Fig.~\ref{fig_results} shows multi-cell tracking performance of our method on multiple datasets. For the consideration of clarity, only a portion of field of view is selected and enlarged. Different kind of cells have different morphology. We track trajectories of cells and get each cell segmentation. Fine segmentation results on highly dense cell population is shown as in Fig.~\ref{fig_results} (a). As shown in Fig.~\ref{fig_results} (c), cells can be segmented accurately even when partly disappeared. Cells can be segmented accurately when their gray level is similar to that of background.

\begin{figure}
    \centering
    \includegraphics[width=0.9\linewidth]{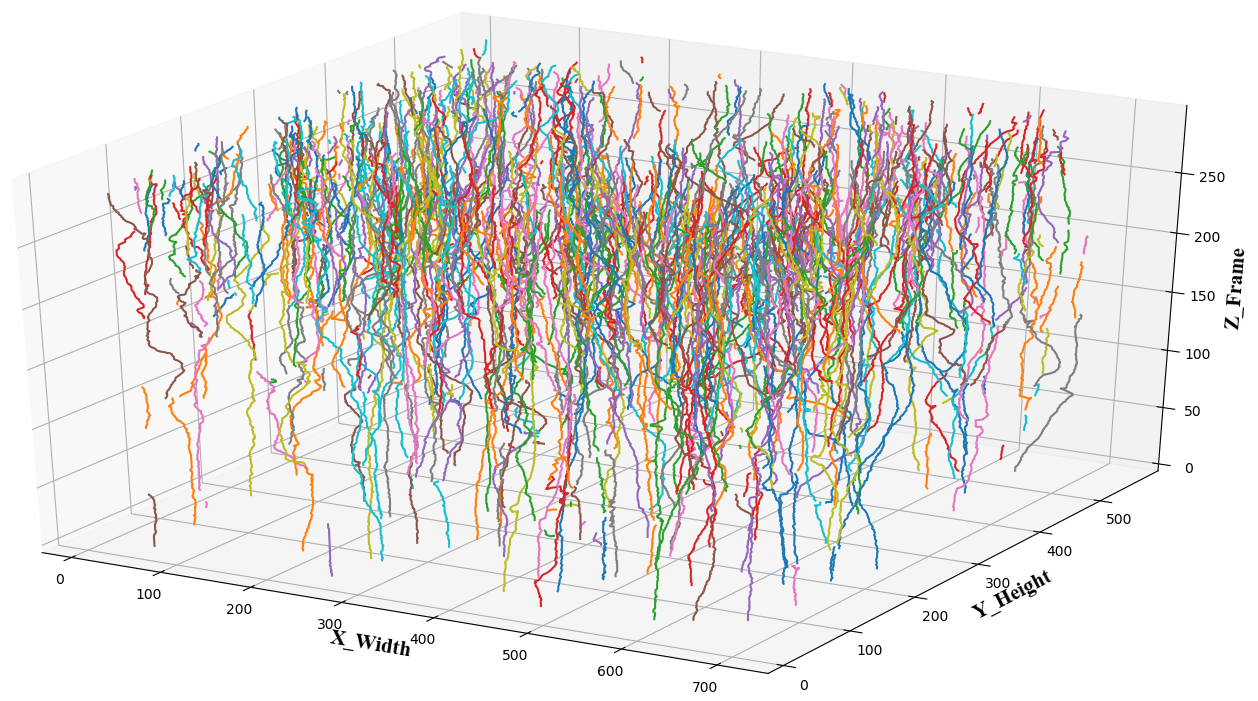}
    \caption{Cell spatio-temporal trajectories of Phc-PSC.}
    \label{fig_trajectory}
\end{figure}

Fig.~\ref{fig_trajectory} shows cell spatio-temporal trajectories of our method on dataset Phc-PSC. It shows trajectories of all cells, as well as evolution of cell lineage. Cells will undergo mitosis over time or leave the field of view. As the frame number increases, cell trajectories become dense.

\section{Conclusion}
We propose a multi-cell tracking framework, which jointly use detection and segmentation. Cell centroid detection is conducted using a UNet with multi-frame input images. Detection of mitotic cells is improved without increasing much network overhead, and therefore improve the detection performance of mitosis event by our mitosis detection algorithm. With our method, normal cells newly entering the field of view can be distinguished from newborn daughter cells. Another UNet is utilized to acquire primary cell segmentation. Fine segmentation is conducted to separate each cell individually, which jointly use primary tracking results of cell centroid detection and primary cell segmentation results.

Evaluations are conducted to compare our method with other methods with datasets in Cell Tracking Challenge. Due to jointly use detection and segmentation, our method performs excellent and achieves a new state-of-the-art performance on dataset Fluo-Hela.

Performance on some datasets is still not very ideal. In future works, fine segmentation will be further optimized, and more image information will be used for more accurately segmentation and tracking.

%
% ---- Bibliography ----
%
% BibTeX users should specify bibliography style 'splncs04'.
% References will then be sorted and formatted in the correct style.
%
% \bibliographystyle{splncs04}
% \bibliography{mybibliography}
%

\end{document}